\documentclass[10pt,twocolumn,letterpaper]{article}

\usepackage[algorithms]{wacv}              % To produce the CAMERA-READY version
\usepackage{graphicx}
\usepackage{amsthm}
\usepackage{amsmath}
\usepackage{amssymb}
\usepackage{booktabs}
\usepackage{multirow}
\usepackage{multicol}
\usepackage{times}
\usepackage{soul}
\usepackage{url}
\usepackage{caption}
\usepackage{enumitem}
\usepackage[table]{xcolor}
\usepackage{paralist}
\usepackage{svg}
\usepackage[accsupp]{axessibility}

\definecolor{bluegray}{rgb}{0.4, 0.6, 0.8}
\definecolor{orange}{HTML}{C55A11}
\usepackage[pagebackref,breaklinks,colorlinks,citecolor=bluegray]{hyperref}

\usepackage[capitalize]{cleveref}
\crefname{section}{Sec.}{Secs.}
\Crefname{section}{Section}{Sections}
\Crefname{table}{Table}{Tables}
\crefname{table}{Tab.}{Tabs.}

\def\wacvPaperID{***} % *** Enter the WACV Paper ID here
\def\confName{WACV}
\def\confYear{2025}

\begin{document}

% \title{\LaTeX\ Author Guidelines for \confName~Proceedings}

\title{\textsc{@Bench}: Benchmarking Vision-Language Models for Human-centered Assistive Technology}

\author{
Xin Jiang$^{1,3,*}$,
~~Junwei Zheng$^{1,*}$,
~~Ruiping Liu$^1$,
~~Jiahang Li$^2$,
~~Jiaming Zhang$^{1,{\dag}}$,\\
~~Sven Matthiesen$^2$,
~~Rainer Stiefelhagen$^1$\\
\normalsize
$^1$CV:HCI, Karlsruhe Institute of Technology,
\normalsize
~~$^2$IPEK, Karlsruhe Institute of Technology,
\normalsize
~~$^3$Li Auto Inc.
}

\twocolumn[{%
\renewcommand\twocolumn[1][]{#1}%
\maketitle
\begin{center}
    \centering
    \captionsetup{type=figure}
    \vskip -3ex
    \includegraphics[width=1.0\textwidth]{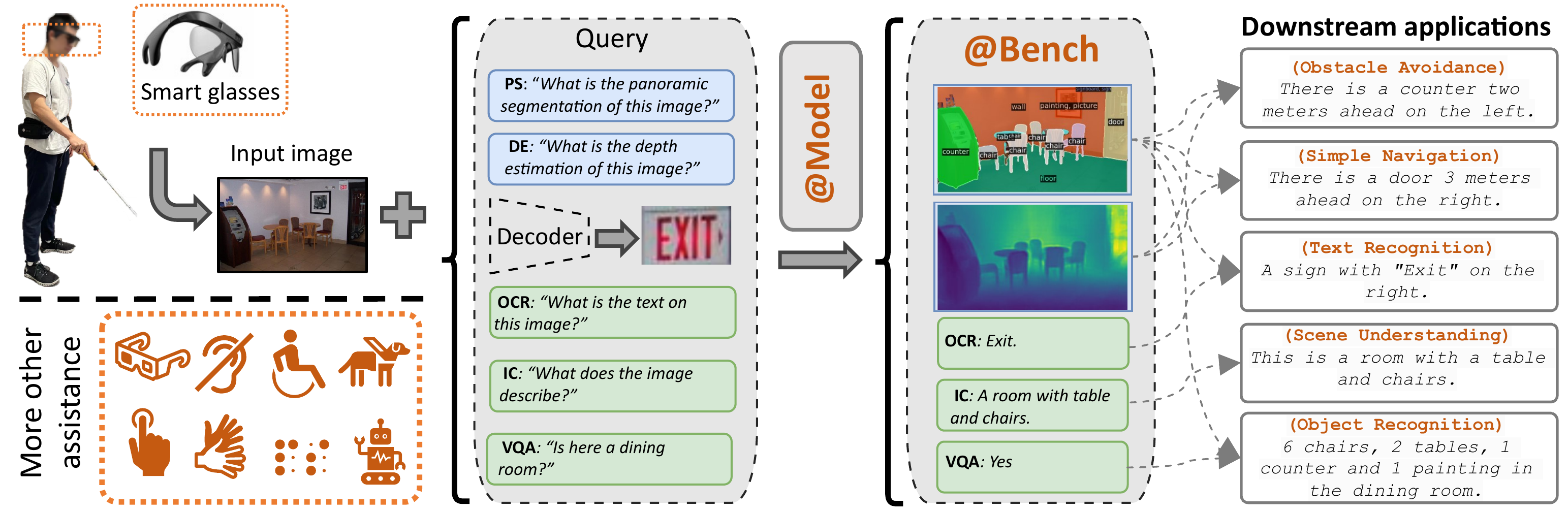}
    \captionof{figure}{
    \textbf{Overview of our Assistive Technology Model (\textsc{\textcolor{orange}{@Model}}) and Benchmark (\textsc{\textcolor{orange}{@Bench}}).}
    \textsc{@Model} can perform vision-language tasks all at once, including: Panoptic Segmentation, Depth Estimation, Image Captioning, Optical Character Recognition and Visual Question Answering. All tasks of \textsc{@Bench} are selected by People with Visual Impairments (PVIs) to evaluate VLMs for AT.} 
    \label{fig:multi_task}
\end{center}%
}]

%%%%%%%%% Footnote
\let\thefootnote\relax\footnotetext{$^*$Equal contribution.}
\let\thefootnote\relax\footnotetext{$^{\dag}$Corresponding author (e-mail: {\tt jiaming.zhang@kit.edu}).}
\let\thefootnote\relax\footnotetext{All codes will be made publicly available at \href{https://junweizheng93.github.io/publications/ATBench/ATBench.html}{ATBench}.}

%%%%%%%%% ABSTRACT
\begin{abstract}
As Vision-Language Models (VLMs) advance, human-centered Assistive Technologies (ATs) for helping People with Visual Impairments (PVIs) are evolving into generalists, capable of performing multiple tasks simultaneously. However, benchmarking VLMs for ATs remains under-explored. 
To bridge this gap, we first create a novel \textbf{AT benchmark (\textsc{@Bench})}. Guided by a pre-design user study with PVIs, our benchmark includes the five most crucial vision-language tasks: \emph{Panoptic Segmentation}, \emph{Depth Estimation}, \emph{Optical Character Recognition (OCR)}, \emph{Image Captioning}, and \emph{Visual Question Answering (VQA)}. Besides, we propose a novel \textbf{AT model (\textsc{@Model})} that addresses all tasks simultaneously and can be expanded to more assistive functions for helping PVIs. Our framework exhibits outstanding performance across tasks by integrating multi-modal information, and it offers PVIs a more comprehensive assistance. Extensive experiments prove the effectiveness and generalizability of our framework.
\end{abstract}

%%%%%%%%% BODY TEXT
\vspace{-3ex}
\section{Introduction}
\label{sec:intro}
Assistive Technologies (ATs) for People with Visual Impairments (PVIs) have witnessed significant advancements in recent years where computer vision and natural language processing play an important role. Previous works~\cite{aladren2014navigation,gurari2018vizwiz,gurari2020captioning,zhang2021trans4trans} focus on providing PVIs with specific and limited functionalities, such as navigation~\cite{aladren2014navigation}, obstacle avoidance~\cite{zhang2021trans4trans}, image captioning~\cite{gurari2020captioning}, \textit{etc.}
However, existing methods face challenges in efficiently processing multiple tasks simultaneously.
Specifically, existing approaches struggle to accurately interpret complex scenes, which is essential to meet the needs of PVIs. Additionally, they often provide less contextually relevant information for scene description.
Recently, the generalist Vision-Language Models (VLMs)~\cite{wang2022ofa,lu2022unified,zou2023generalized} have been proposed and shown great potential in multi-tasking and revolutionizing the next-generation assistive systems. 
Regarding Vision-Language (VL) tasks, these models benefit from the dividend in both computer vision and natural language processing domains, facilitating collaboration between visual and language tasks for a more comprehensive understanding of the surrounding environment.
Nonetheless, benchmarking VLMs for ATs is still under-explored.
The previous Large Language Models (LLMs) and Large Vision-Language Models (LVLMs) benchmarks~\cite{liu2023mmbench,zhong2023agieval} have limitation in two aspects. Firstly, while the benchmarks focus on language-specific and cross-modal tasks, they seldom address pure vision-specific tasks, which are crucial in the context of visual impairments. Secondly, these benchmarks tend to prioritize general applicability over the specific needs of individuals with visual impairments.
Therefore, we raise the research question: \textbf{\textit{Are VLMs ready for empowering assistive technology for helping People with Visual Impairments (PVIs)?}} 

To answer this question, 
we propose a new VL-based {AT benchmark (\textsc{@Bench})} as a platform for evaluating VLMs for visually impaired assistance. 
To involve the target group in shaping the benchmark, we conduct a number of questionnaires via a human-in-the-loop process with seven participants who are blind or have low vision, so as to understand the practical demands of PVIs. Based on the feedback of the user study, we introduce five tasks ranked by blind users according to the level of interest, frequency of usage, and level of importance. As presented in Fig.~\ref{fig:multi_task}, the most assistance-related tasks include: \emph{Panoptic Segmentation (PS)}, \emph{Depth Estimation (DE)}, \emph{Optical Character Recognition (OCR)}, \emph{Image Captioning (IC)}, and \emph{Visual Question Answering (VQA)}.
Beyond the range of tasks, performance and efficiency stand as crucial considerations in AT for PVIs. Therefore, we introduce an evaluation framework for assessing the trade-off between efficiency and performance in generalist VLMs.

With this benchmark in place, we propose a novel {AT model (\textsc{@Model})} that use \textit{task-specific prompt} to combine these 5 uni-modal or cross-modal tasks and realize the paradigm of multi-task training. Thanks to this, \textsc{@Model} can use one suit of parameters to implement multiple tasks. It is crucial to significantly reduce the number of parameters, and it will be possible to deploy {one} model and {one} suit of weights on the portable device for PVIs.

To summarize, we present the following contributions:

\begin{compactitem}
\item[(1)] \textbf{Human-in-the-loop User Study.} As a part of PVIs-specific design, it is necessary to investigate the needs of the target group. We conduct a participatory user study for the sake of understanding the most related tasks, enabling a user-driven design.

\item[(2)] \textbf{Vision-Language Benchmark for Assistive Technology.} We release a new benchmark with five representative VL tasks close to the daily life of PVIs. Other complex functionalities can be derived from these tasks, such as obstacle avoidance. We further evaluate the efficiency-performance trade-off on \textsc{@Bench}.

\item[(3)] \textbf{Generalist Vision-Language Model for Assistive Technology.} We propose a new end-to-end baseline \textsc{@Model} for addressing multiple tasks in \textsc{@Bench} all at once. The model achieves competitive performance compared with state-of-the-art methods. 

\item[(4)] \textbf{One Suit of Weights for All Tasks.} Benefiting from multi-task training, our model is capable of concurrently executing all tasks with a unified set of weights, resulting in a significant reduction in the number of parameters and computational cost.

\end{compactitem}

%%%%%%%%% 
\section{Related Work}
\label{related_work}
\subsection{Assistive Technologies for the Blind}
A common goal in ATs is to develop artificial intelligent systems via vision-language algorithms to help PVIs. 
VizWiz, introduced by Bigham \textit{et al.} in 2010~\cite{bigham2010vizwiz}, presents the first multi-task datasets and artificial intelligence challenges originating from PVIs. These include over 10 tasks, such as VQA~\cite{gurari2018vizwiz}, image captioning~\cite{gurari2020captioning}, object detection~\cite{reynolds2023salient} and object classification~\cite{bafghi2023new}, \textit{etc.} 
It covers various scenes in the daily life of PVIs and provides valuable data for the research of ATs. 
Other task-specific datasets~\cite{tang2023dataset,zhang2023grfb} focus on the recognition of obstacles and tactile paving, further contributing to this field. 
Based on the PVIs-oriented and general datasets, most work used visual model to address the daily challenges encountered by PVIs.
For example, previous works~\cite{aladren2014navigation,wang2017enabling,zhang2021trans4trans,gurari2019vizwiz,zheng2024materobot,liu2023opensu} employ visual tasks such as detection, segmentation and depth estimation to accomplish avoidance, navigation, privacy protection, \textit{etc.}
Some other works have focused on language-modal or cross-modal tasks,  such as OCR~\cite{qiao2020seed,lyu2022maskocr}, image captioning~\cite{gurari2020captioning}, and VQA~\cite{gurari2018vizwiz}. Compared to these existing methods, we conduct a human-in-the-loop study to design a unified multi-modal benchmark for evaluating VLMs for ATs.

\subsection{Generalist Vision-Language Models}
Generalist VLMs have witnessed remarkable advancements, driven by breakthroughs in deep learning techniques~\cite{radford2021learning,wang2022ofa,zou2023generalized}. 
Developing a generalist model for multiple tasks poses unique challenges due to the heterogeneous inputs and outputs, including RGB images, depth maps,
binary masks, bounding boxes, language, etc. 
Previous methods MetaLM~\cite{hao2022language} and PaLI~\cite{chen2022pali} use language models as general-purpose interfaces to various foundation models. 
GLIPv2~\cite{zhang2022glipv2} unifies both localization and VL understanding tasks as grounded vision-language tasks. 
OFA~\cite{wang2022ofa} and Unified-IO~\cite{lu2022unified} introduce a Seq2Seq framework for the unification of I/O, architectures, tasks, and modalities. 
X-Decoder~\cite{zou2023generalized} can predict pixel-level segmentation and language tokens through a generalized decoding model.
Uni-Perceiver v2~\cite{li2023uni} formulates different tasks as a unified maximum likelihood estimation problem without any task-specific fine-tuning. However, these methods are either unable to perform multi-task training, or they focus too much on language-modal tasks or cross-modal tasks, while ignoring vision-modal tasks that are important for PVIs, such as segmentation~\cite{zheng2025ops} and depth estimation~\cite{li2024binsformer}. Therefore, we propose a new method that can comprehensively consider and balance multiple assistance-related uni-modal and cross-modal tasks, and can use one suit of weights for all tasks.

\subsection{Benchmarks for Vision-Language Models}
To evaluate vision-language systems, Zhou~\textit{et al.}~\cite{zhou2022vlue} propose a multi-task multi-dimension benchmark for Vision Language Pretraining (VLP) models. Su~\textit{et al.}~\cite{su2021gem} introduce the GEM benchmark, a multi-modal benchmark that focuses on both image-language tasks and video-language tasks. Recently, many LVLMs benchmarks ~\cite{liu2023mmbench,zhong2023agieval} have emerged to more comprehensively evaluate the fine-grained capabilities of models.
However, a vision-language benchmark for ATs and PVIs is lacking in the literature. To bridge this gap, we introduce \textsc{@Bench} , a benchmark that includes realistic multi-modal tasks closely relevant to the daily lives of PVIs. 
Therefore, \textsc{@Bench} is designed to serve as a fundamental benchmark for evaluating VLMs in the realm of ATs.

%%%%%%%%% 
\section{\textsc{@Bench}: Assistive Technology Benchmark}

\begin{table*}[t]
\centering
\renewcommand\arraystretch{1.4}
\setlength{\tabcolsep}{8pt}
\resizebox{\linewidth}{!}
{%
\begin{tabular}{lll|cccc}
\toprule[1pt]
\multirow{2}{*}{} & \multirow{2}{*}{Function}  & \multirow{2}{*}{Related Task} & Level of Interest                      & Frequency of Usage              & Importance                                  & \multirow{2}{*}{Total} \\
                           &     &                          & 1 (low)– 5 (high) & 1 (low) – 5 (high) & 1 (low) – 5 (high) &                        \\\hline
1 & Obstacle Avoidance   & Panoptic Segmentation         & 3.00   & 2.71                             & 2.43                                            & \cellcolor[HTML]{b7e4c7}08.14                      \\
2 & Indoor Distance Estimation & Depth Estimation              & 3.14                                      & 2.43                             & 2.43                                           & \cellcolor[HTML]{d8f3dc}08.00                      \\
3 & Object Recognition         & Panoptic Segmentation         & 3.86                                        & 3.71                               &  4.00                                           & \cellcolor[HTML]{40916c}11.57                        \\
4 & Object Location            & Panoptic Segmentation              & 3.29                                         & 3.14                             & 3.29                                            & \cellcolor[HTML]{74c69d}09.72                       \\
5 & Text Recognition           & OCR             & 4.57                                        & 4.43                               & 4.43                                            &  \cellcolor[HTML]{40916c}13.43                      \\
6 & Surroundings Understanding        & Image Captioning              & 3.43                                      & 2.86                               &  2.71                                           & \cellcolor[HTML]{52b788}09.00                      \\
7 & Scene Recognition          & Scene Recognition             & 2.14                                       & 1.71                               & 1.86                                            & \cellcolor{gray!25}05.71                       \\
8 & Visual Q\&A  & Visual Question Answering     & 3.57                                       & 3.43                               & 3.43                                            & \cellcolor[HTML]{52b788}10.43                       \\ 
\bottomrule[1pt]
\end{tabular}%
}
\caption{\textbf{Quantitative result of the user study.} Potential functions can be achieved via related tasks, which are listed in the questionnaires for the user study. Note: all scores are averages across 7 participants.}
% \vspace{-1em}
\label{survey}
\end{table*}

\textsc{@Bench} is a pioneering multi-modal benchmark tailored specifically for the domain of Assistive Technology (AT). The primary target of \textsc{@Bench} is to establish a comprehensive and standardized evaluation platform for vision-language models in the context of helping PVIs.

In this section, we introduce the detail of the user study (Sec.~\ref{sec:user_study}), which helps us identify important tasks that are closely related to PVIs. We then provide an overview (Sec.~\ref{sec:assistive_tasks}) of the tasks, datasets and metrics encompassed within the \textsc{@Bench} framework. Subsequently, we introduce the vital dimension for assessing the performance of VLMs for ATs: efficiency-performance trade-off (Sec.~\ref{sec:trade_off}).

\begin{table}[]
\centering
\renewcommand\arraystretch{1.5}
\setlength{\tabcolsep}{6pt}
\resizebox{\linewidth}{!}{%
\begin{tabular}{ll|lll|l}
\toprule[1pt]
Task                          & Dataset     & \#Train & \#Val & \#Test & Metric\\
\midrule[0.5pt]
PS         & ADE20K         &25,574         &2,000       &2,000        & PQ\\
DE              & NYU v2  &24,230        &654      &654        & RMSE \\ 
OCR&  MJ, ST, 6 OCR          &15,895,356    &7,507     &7,507      & Accuracy\\
IC              & VizWiz\_Cap  &23,431         &7,750       &8,000        & BLEU-1/CIDEr \\
VQA     & VizWiz\_VQA&20,523        &4,319       &8,000      & Accuracy\\ \midrule
$\sum$ & @\textsc{Bench} & 15,989,114 & 22230 & 26161 & \\
\bottomrule[1pt]
\end{tabular}%
}
\caption{\textbf{Statistic of pre-selected tasks and datasets in \textsc{@Bench}}. Note that some datasets do not have a \textit{test} subset, so use \textit{val} subset for evaluation. The 6 OCR datasets are IC13, IC15, IIIT5K, SVT, SVTP and CUTE.}
% \vspace{-1em}
\label{benchmark}
\end{table}

\subsection{User-centered Study}\label{sec:user_study}
\noindent \textbf{Organization.} To build the AT benchmark, we conducted a pre-study with 2 accessibility experts to develop a reasonable questionnaire. Based on their suggestions and multiple discussions, we further organized a user-centered study with 7 participants who are blind or have low vision. The goal was to identify which vision-language tasks are beneficial for the target group and meet their requirements. 

\noindent\textbf{Questionnaire.} 
To enhance the rationale of our benchmark design, we conducted a questionnaire session with the participants. In this questionnaire, we presented 8 functions relevant to the daily lives of PVIs in Table~\ref{survey}: (1) \textit{obstacle avoidance}, (2) \textit{indoor distance estimation}, (3) \textit{object recognition}, (4) \textit{object location}, (5) \textit{text recognition}, (6) \textit{surroundings understanding}, (7) \textit{scene recognition} and (8) \textit{visual Q\&A for surroundings}. At the begining of the user study, we first explained the specific concepts of different functions and related usage scenarios to the participants. Then, they were asked to rate each function based on 3 criteria: (1) level of interest, (2) frequency of usage, and (3) importance in their daily life from 1 (lowest) to 5 (highest). In addition, each function has and can be performed by using a corresponding uni-modal or cross-modal task. 

\noindent\textbf{Quantitative Result.} Each participant rated each function from 1 to 5 based on these three criteria. After collecting the scores of each participant, we averaged the scores of the 7 participants to represent the score of the function, and finally used the total score of the 3 criteria as the score of each function. And we use the total score as a basis to select relevant functions and tasks.
The ratings were then aggregated and summarized, as illustrated in Table~\ref{survey}. 
Functions 5 and 3 have the highest scores. PVIs think text recognition and object recognition are the most important in their daily life. Function 7 has the lowest score and has a large gap with other functions. At the same time, functions 1, 2, 4, 6, and 8 have similar scores. Therefore, we did not consider the tasks corresponding to function 7 in the benchmark, and retained the tasks related to all the remaining functions. {According to the study, the selected functions are: panoptic segmentation, depth estimation, OCR, image captioning and VQA.} 

% xxxxxxxxxxxxx

\subsection{Assistive Tasks}\label{sec:assistive_tasks}
Guided by the user study, \textsc{@Bench} contains 5 tasks that are extremely relevant to the daily lives of PVIs.  
We give an overview of all tasks and the corresponding datasets in Table~\ref{benchmark}, and describe the details as follows:

\noindent\textbf{Panoptic Segmentation.} It is the task that combines both semantic segmentation and instance segmentation, aiming to simultaneously recognize all object instances in an image and segment them by category, helping blind people perceive the surroundings more accurately. We opt to ADE20K ~\cite{zhou2019semantic}, which contains more than 27K images spanning 365 different scenes with totally 150 semantic categories, including indoor, outdoor, urban, rural, \textit{etc.}, basically covering almost all daily life scenes and common objects of PVIs. 
And we use Panoptic Quality (PQ) ~\cite{kirillov2019panoptic} to measure the performance.

\noindent\textbf{Depth Estimation.} It is the task of measuring the distance of each pixel relative to the user's camera. According to user study, the majority of PVIs spend most of their time indoors. Therefore, we choose NYU v2~\cite{Silberman:ECCV12}, which includes indoor scenes. We evaluate with the RMSE metric. 

\noindent\textbf{Optical Character Recognition.} It is the conversion of images of typed, handwritten or printed text into machine-encoded text. We know text recognition is the most important function for the PVIs from user study, so we select two widely-used, large synthetic datasets for training: MJSynth (MJ)~\cite{jaderberg2014synthetic} and SynthText (ST)~\cite{jaderberg2014synthetic}. We evaluate recognition accuracy on 6 datasets, covering various text scenarios: ICDAR 2013 (IC13)~\cite{karatzas2013icdar}, ICDAR 2015 (IC15)~\cite{karatzas2015icdar}, IIIT5K-Words (IIIT5K)~\cite{mishra2012scene}, Street View Text (SVT)~\cite{wang2011end}, Street View Text-Perspective (SVTP)~\cite{phan2013recognizing}, and CUTE80 (CUTE)~\cite{risnumawan2014robust}. 

\noindent\textbf{Image Captioning.} 
It is a challenging task that involves generating human-like and coherent natural language descriptions for images. The task is to comprehend visual content and express in natural language that is descriptive and contextually relevant, which allows PVIs to have an overall understanding of their surroundings.
To better align with the daily experiences of PVIs, we opted for the VizWiz\_Cap~\cite{gurari2020captioning}, collected from the perspective of PVIs. We use BLEU-1~\cite{papineni2002bleu} and CIDEr~\cite{vedantam2015cider} as evaluation metrics.

\noindent\textbf{Visual Question Answering.} It requires the model to take as input an image and a free-form, open-ended, natural language question. It produces or selects a natural language answer as output~\cite{zhou2022vlue}. In our work, we found that PVIs expressed significant interest in this task. By posing questions, they can experience and comprehend the unseen world around them, providing them with a novel and enriching experience. We select VizWiz\_VQA~\cite{gurari2018vizwiz} and use the publicly released VizWiz\_VQA evaluation scripts in \textsc{@Bench}.

\subsection{Efficiency-Performance Trade-off}\label{sec:trade_off}
In designing models for assistive systems, an optimal balance between efficiency and performance is essential. 
Normally, the performance of VLMs can be easily measured by task-specific metrics. For efficiency, there are few common choices: the number of parameters, FLOPs and inference time. To compare with previous methods, we evaluate efficiency through the number of parameters.

In sum, \textsc{@Bench} is designed specifically for multi-modal, multi-task scenarios and tailored to assistive systems for PVIs. All tasks are closely tied to the needs of PVIs community. \textsc{@Bench} not only prioritizes performance but also places emphasis on efficiency-performance trade-off. We aspire for this benchmark to serve as a cornerstone for researchers within PVIs assistance community, encouraging exploration of multi-modal models' applications in ATs.

%%%%%%%%% 
\section{\textsc{@Model}: Assistive Technology Model}
\label{baseline}
\begin{figure}[!t]
\centering
\includegraphics[width=1.0\linewidth]{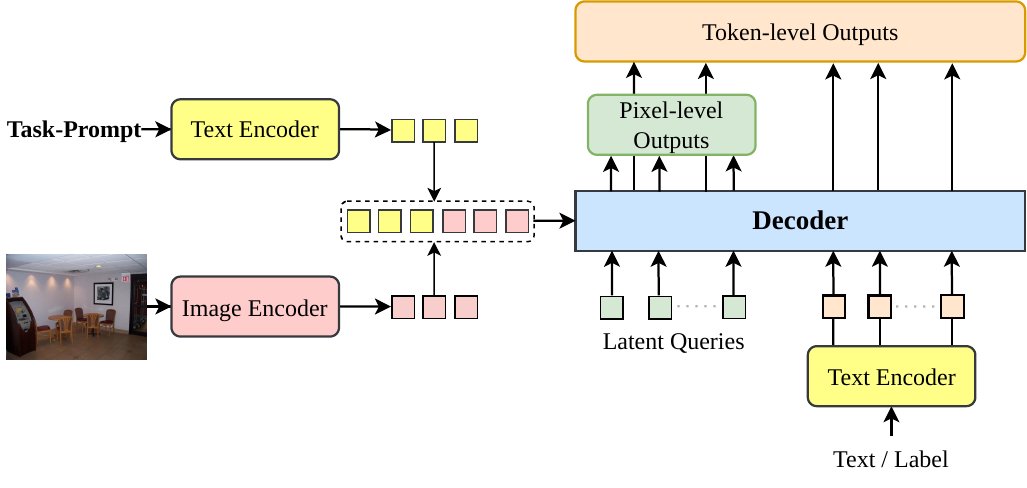}
\caption{\textbf{Overall architecture of \textsc{@Model}}. We propose task-based prompts to unify inputs and perform different tasks all at once.}
% \vspace{-1em}
\label{fig:model_arch}
\end{figure}

Based on X-Decoder, we propose \textsc{@Model}, the first generalist model to support all these assistance-related vision-language tasks. As show in Fig.~\ref{fig:model_arch}, the overall model is built on top of a image encoder for extracting image features, two text encoders that share parameters for extracting text features, and a transformer decoder with generic latent queries and textual queries. 
\textsc{@Model} has two types of output: (1) pixel-level output for dense prediction, such as panoptic segmentation and dense estimation and (2) token-level output for a diverse set of language-related vision tasks, such VQA, image captioning and OCR. 

\begin{figure}[!t]
\centering
\includegraphics[width=0.5\textwidth]{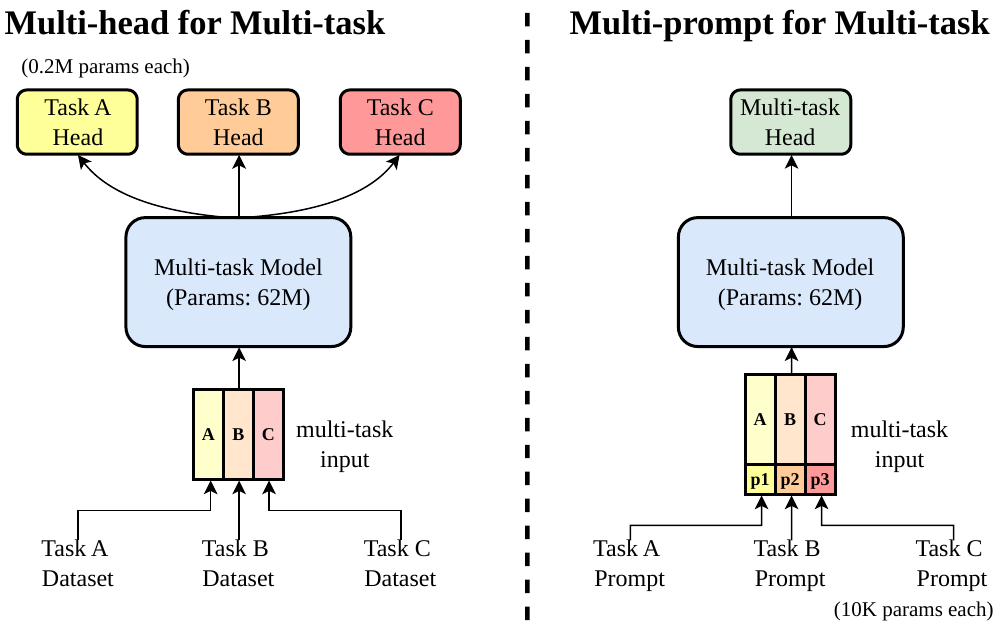}
\caption{\textbf{Paradigms of multi-task methods.} Our \textsc{@Model} incorporates task-specific prompts that effectively unify tasks all at once and with almost no additional parameters.
}
% \vspace{-1em}
\label{fig:paradigm}
\end{figure}

\noindent\textbf{Unified Multi-task Architecture.} 
X-Decoder~\cite{zou2023generalized} includes only two tasks in \textsc{@Bench},  panoptic segmentation and image captioning. To include token-level output like VQA or OCR, X-Decoder requires a specific head for each task. This paradigm (in Fig~\ref{fig:paradigm}) will make the structure of the entire model very bloated, which is a major drawback for portable assistance systems. In contrast,  
we use task-specific prompt to build a unified input paradigm ``image + prompt'' as shown in Fig.~\ref{fig:paradigm}. Compared with multi-head output, the benefits are three-fold: (1) Unifying input forms for different tasks. For example, it can unify the inputs of VQA (image and question text) and segmentation (image) in the manner of ``image + prompt''. (2) Enabling the model to distinguish different tasks, extract corresponding features in early phase. (3) Reducing the number of parameters.
We design a corresponding prompt shown in  Fig.~\ref{fig:multi_task} for each task. For VQA, we use its own questions directly as the prompt. In Sec.~\ref{sec:unified-input}, we analyze the performance and efficiency advantages of such a design. 

\noindent\textbf{Character-based Tokenizer with Limited Vocabulary for OCR.} 
In the text encoders of \textsc{@Model}, we use pretrained CLIP~\cite{radford2021learning} subword-based tokenizer with a vocabulary containing approximately 50,000 subwords as default.  
But during language-related tasks training, we need to consider a problem, there is a mismatch between dataset's vocabulary space and model's prediction vocabulary space.
For example, the dataset of captioning contains a rich vocabulary that is comparable to the vocabulary that can be predicted by the model. But for OCR, we have some observations: (1) In the English OCR datasets in \textsc{@Bench}, an image usually contains only one pure text. (2) The texts basically use 26 English letters and 10 numbers. (3) If each text is divided by a single character, the length is basically less than 15. When we use the default subword-based tokenizer and complete vocabulary, we found that the training effect is unsatisfactory. Subword-based tokenizer with a large vocabulary that can provide semantic information is not effective enough for OCR. The model only needs to recognize the text but not the representations of the text. A character-based tokenizer can bring a much smaller vocabulary and relieve this mismatch. More details locate in the supplementary material. The respective experiment is presented in Sec.~\ref{sec:ch_OCR}. 

\begin{figure*}[!t]
  \centering
   \includegraphics[width=1.0\linewidth]{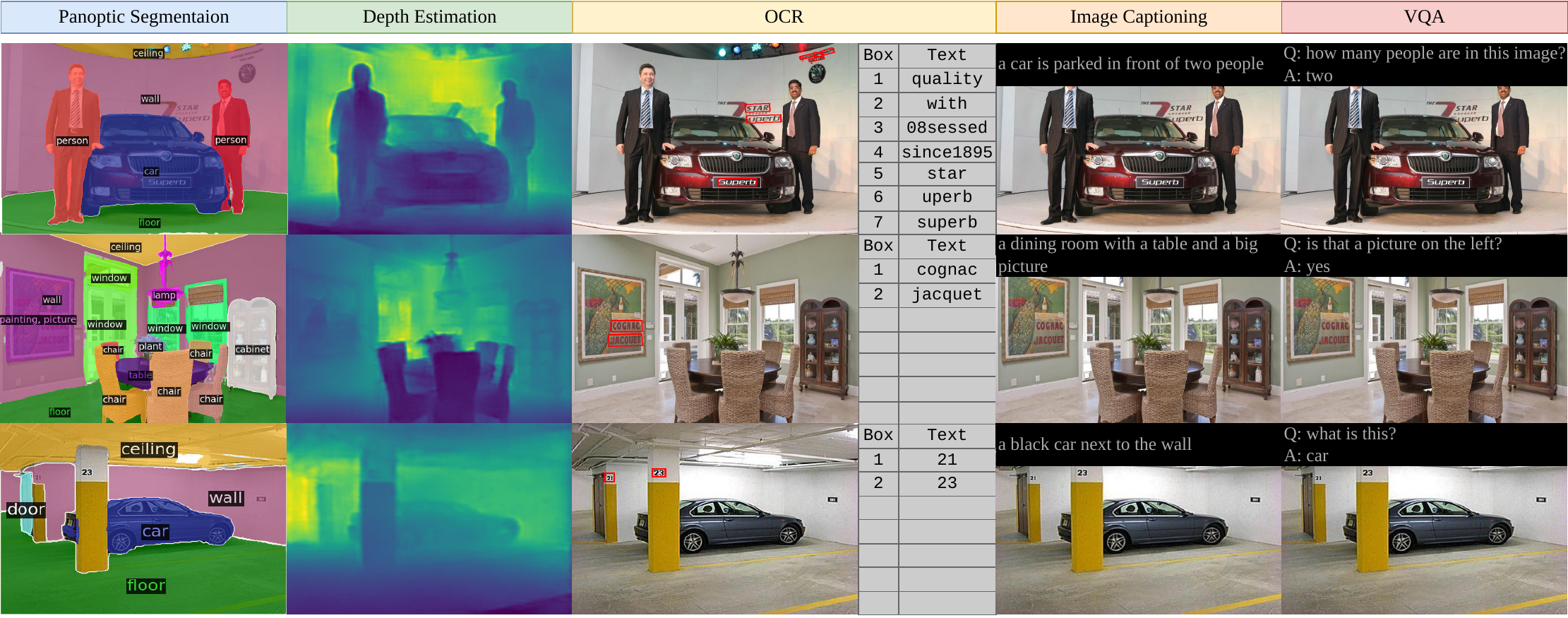}
   \caption{\textbf{Examples of multi-task training results on 5 tasks.} Given one image as input our \textsc{@Model} can output all predictions.
   }
   % \vspace{-1em}
   \label{fig:example_show}
\end{figure*}

%%%%%%%%% 
\section{Experiments}
\label{experiments}
This section provides experimental results and analyses to demonstrate the effectiveness of our proposed model. Implementation details are in the supplementary material.

\begin{table}[t]
% \vspace{2ex}
\renewcommand\arraystretch{1.9}
\setlength{\tabcolsep}{2pt}
\resizebox{\linewidth}{!}{%
\begin{tabular}{l|c|c|c|cc|c|c}
\hline
\multirow{3}{*}{Method} & \underline{\emph{PS}}  & \multicolumn{1}{c|}{\underline{\emph{DE}}}                                           & \multicolumn{1}{c|}{\underline{\emph{OCR}}}                      & \multicolumn{2}{c|}{\underline{\emph{IC}}} & \underline{\emph{VQA}}          & \multirow{3}{*}{\#Params} \\ 
                        & ADE-150               & \multicolumn{1}{c|}{NYU-V2}                                                    & 6 Datasets avg& \multicolumn{2}{c|}{VizWiz\_Cap} & VizWiz\_VQA &         \\
                        & PQ                    & RMSE $\downarrow$ & \multicolumn{1}{c|}{Acc($\%$)}                       & B@1 & CIDEr & Acc($\%$)        &         \\ \hline
Unified-IO (S)$^{\dag} $       & --                    & 0.649          & --     & $\star$    & $\star$     & 42.4       & 71M     \\
Unified-IO (B)$^{\dag} $         & --                    & 0.469         & --     & $\star$    & $\star$     & 45.8       & 241M     \\
Unified-IO (L)$^{\dag} $         & --                    & 0.402          & --     & $\star$    & $\star$     & 47.7       & 776M     \\
X-Decoder (T)$^{\dag} $           & 41.6                     &--              & --       & $\star$      & $\star$     & --          &164M         \\  
GIT$^{\dag} $ &--                    & --          & --      & $\star$    & 113.1     & 68.0       & 0.7B     \\ 
PaLI$^{\dag} $ &--                    & --          & --      & $\star$    & 117.2     & 67.5       & 3.0B     \\ \hline
\rowcolor{gray!15} Ours                    &38.5                     &0.425          &80.1    &61.0     &52.5       &53.7            &62M         \\ \hline
\end{tabular}%
}
\caption{\textbf{Comparison of multi-task training \textsc{@Model} and other generalist models.} We report the multi-task training results without any pre-training and task-specific fine-tuning. Note: GIT and PaLI are LVLMs. ``$\star$'' denotes the model has the capability for the task but does not have number reported. ``--'' means the model does not have the ability for the specific task. ``${\dag}$'' means the model uses pre-trained weights for training. (B@1 = BLEU-1)}
% \vspace{-1em}
\label{exp:multi_task}
\end{table}

\subsection{Comparison with Existing Generalist Models}
In Table~\ref{exp:multi_task}, we compare the performances of \textsc{@Model} and some other generalist models on the tasks in \textsc{@Bench}. 
\textsc{@Model} is the first generalist model to support assistance-related vision-language tasks and can achieve superior results. However, (1) most generalist models only include a few specific tasks and do not include all tasks in \textsc{@Bench}, especially OCR task. And (2) most generalist models only report fine-tuned results and few methods report their results on assistance-related datasets. Therefore, we only compare a few related methods.
Since generalist models aim to process different tasks with shared architecture and parameters, some generalist models only report results with task-specific fine-tuning, \textit{e.g.}, X-Decoder, GIT~\cite{wang2022git} and PaLI, this fine-tuning will lose the general modeling ability, so we report the numbers without any task-specific adaptation in the manner of multi-task training. 
Nonetheless, with similar number of parameters, \textsc{@Model} can outperform Unified-IO (S) in depth estimation and VQA by $0.224$ and $11.3\%$, respectively, and even better than the Unified-IO (B). On other tasks, since many methods are pre-trained, fine-tuned, and have a larger number of parameters, there is still a certain gap between \textsc{@Model}  and these methods.
The visualization of some multi-task training results is presented in Fig.~\ref{fig:example_show}.

\begin{table*}[]
\renewcommand\arraystretch{1.9}
\setlength{\tabcolsep}{2pt}
\resizebox{\linewidth}{!}{%
\begin{tabular}{l|c||l|c||l|ccccccc||l|cc||l|cc}
\hline
\multirow{3}{*}{Method} & \underline{\emph{PS}}      & \multirow{3}{*}{Method} & \underline{\emph{DE}}        & \multirow{3}{*}{Method} & \multicolumn{7}{c||}{\underline{\emph{OCR}}}                     & \multirow{3}{*}{Method} & \multicolumn{2}{c||}{\underline{\emph{IC}}} & \multirow{3}{*}{Method} & \multicolumn{2}{c}{\underline{\emph{VQA}}}         \\
                        & ADE-150 &                         & NYU-V2 &                         & IC13 & IC15 & SVT & IIIT5K & SVTP & CUTE & avg &                         & \multicolumn{2}{c||}{VizWiz\_Cap} &                         & \multicolumn{2}{c}{VizWiz\_VQA} \\
                        & PQ      &                         & RMSE$\downarrow$   &                         & \multicolumn{7}{c||}{Acc ($\%$)}                       &                         &  B@1            & CIDEr            &                         & \multicolumn{2}{c}{Acc ($\%$)}         \\ \hline
MaskFormer$^{\dag} $ (45M)      &34.7         & GLP$^{\dag} $ (62M)                                     &0.344         & ASTER                      & 91.8 &76.1 & 89.5 & 93.4   & 78.5 & 79.5 & 86.7     & VizWiz\_Cap $^{\dag} $                                       &62.1               &48.2                  & VizWiz\_VQA                                       &\multicolumn{2}{c}{47.5}               \\
Mask2Former$^{\dag} $ (44M)   &39.7         & DPT*$^{\dag} $ (123M)                                         &0.357         & SEED                                        & 92.8 & 80.0 & 89.6 & 93.8   & 81.4 & 83.6 & 88.3   &  AoANet                                           &65.9               &59.7                    & S-VQA                                 & \multicolumn{2}{c}{51.6}               \\
kMaX-DeepLab$^{\dag} $ (57M)        &41.5             & BTS$^{\dag} $ (47M)                                       &0.392         & MaskOCR$^{\dag} $ (97M)                      & 98.1 & 87.3 & 94.7 & 95.8   & 89.9 & 89.2 & 93.1            %& IBM\_VizWiz                                  &*               &66.8
&-- & -- & --& CS-VQA                               &\multicolumn{2}{c}{53.2}                \\ \hline
\rowcolor{gray!15} Ours (62M)                    &39.2         & Ours (62M)                                         &0.386        & Ours (62M)                                         & 97.1     & 84.4     & 92.6    & 90.3       & 88.7     & 93.1     & 90.0       & Ours (62M)                                         &60.0            &45.1                 & Ours (62M)                                         &\multicolumn{2}{c}{49.1}             \\ \hline
\end{tabular}%
}
\caption{\textbf{Comparison of single-task training \textsc{@Model} and specialized SoTA models.} Note: ``model (\#params)'' donates the number of parameters of the model. DPT* is trained with an extra dataset. 
}
% \vspace{-1em}
\label{exp:single_task}
\end{table*}

\subsection{Comparison with Specialized SoTA Models}
Due to the scarcity of existing VLMs capable of encompassing all five tasks of our \textsc{@Bench}, to comprehensively showcase the effectiveness of our model, we compare our multi-task model with previous single-task SoTA models in Table~\ref{exp:single_task}. The representative works for each task are: MaskFormer~\cite{cheng2021per}, Mask2Former~\cite{cheng2022masked} and kMaX-DeepLab~\cite{yu2022k} for panoptic segmentation; BTS~\cite{lee2019big}, DPT~\cite{ranftl2021vision} and GLP~\cite{kim2022global} for depth estimation; ASTER~\cite{shi2018aster}, SEED~\cite{qiao2020seed} and MaskOCR~\cite{lyu2022maskocr} for OCR; VizWiz\_Cap, AoANet~\cite{huang2019attention} for captioning, VizWiz\_VQA, S-VQA~\cite{kazemi2017show} and CS-VQA~\cite{jayaram2021cross} for VQA. 
Note that all these specialized methods are tailored for their respective specific tasks, while \textsc{@Model} is proposed to cover all these tasks. 
\textsc{@Model} can achieve, and in some cases, even surpass the single-task SoTA model on multiple tasks.
For example, \textsc{@Model} outperforms the MaskFormer ($+ 4.5\%$) and gets comparable results with pre-trained models, such as Mask2Former and kMaX-DeepLab. Even for highly competitive OCR tasks, \textsc{@Model} can outperform non-pretrained SoTA models, ASTER and SEED, by $3.3\%$ and $1.7\%$, respectively.
These results show that our \textsc{@Model} significantly bridges the performance gap between generalist models and strong baselines.

\subsection{Multi-task Training \textit{v.s.} Single-task Training}

\begin{figure}[t]
\centering
\includegraphics[width=1.0\linewidth]{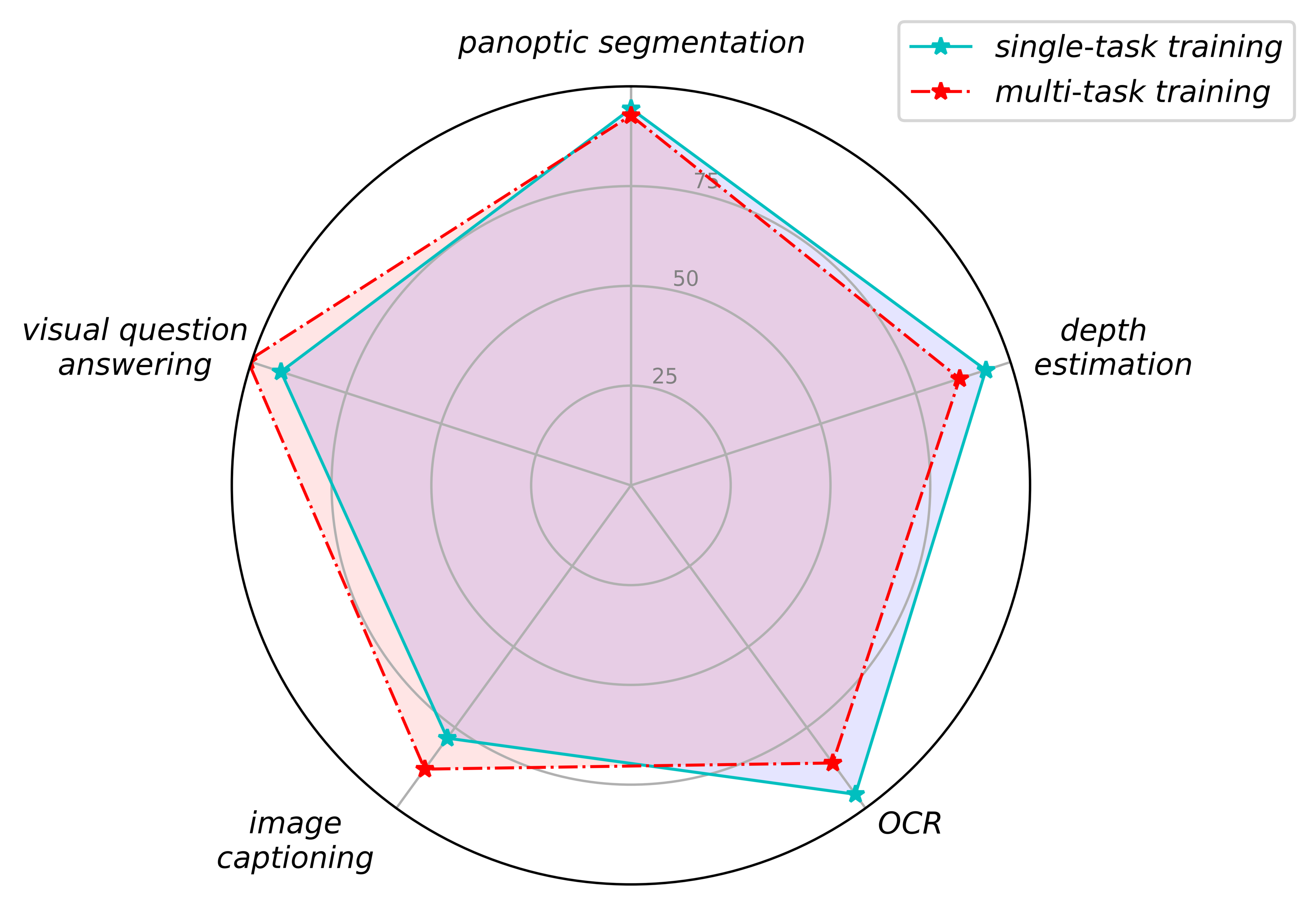}
\caption{\textbf{Single-task and multi-task training performance} (relative) against the specialized SoTA models on different tasks}
% \vspace{-1em}
\label{fig:radar}
\end{figure}

As depicted in Fig.~\ref{fig:radar}, \textsc{@Model} performs less effectively in segmentation, depth estimation, and OCR under multi-task training, while demonstrating superior performance in captioning and VQA. We analyze that due to the huge OCR datasets, it is difficult to balance various tasks during multi-task training, resulting in a decline in OCR performance. For captioning and VQA, these two tasks are related to the scene understanding, they can promote each other during joint training. Furthermore, panoptic segmentation can increase the scene perception ability. See Appendix.  for more analysis.

\begin{table*}[]
\renewcommand\arraystretch{2}
\setlength{\tabcolsep}{4pt}
\resizebox{\linewidth}{!}{%
\begin{tabular}{lc|c|c|ccccccc|cc|c|l}
\hline
\multirow{3}{*}{Method}   & \multirow{3}{*}{Type}                  & \multicolumn{1}{c|}{\underline{\emph{PS}}} & \multicolumn{1}{c|}{\underline{\emph{DE}}}                                     & \multicolumn{7}{c|}{\underline{\emph{OCR}}}                         & \multicolumn{2}{c|}{\underline{\emph{IC}}}         & \underline{\emph{VQA}}                & \multicolumn{1}{c}{\multirow{3}{*}{\#Params}} \\
                          &                                        & \multicolumn{1}{c|}{ADE-150}               & \multicolumn{1}{c|}{NYU-V2}                                               & IC13 & IC15 & SVT  & IIIT5K & SVTP & CUTE & avg  & \multicolumn{2}{c|}{VizWiz\_Cap}        & VizWiz\_VQA        & \multicolumn{1}{c}{}                          \\
                          &                                        & \multicolumn{1}{c|}{PQ}                    & RMSE $\downarrow$ & \multicolumn{7}{c|}{Acc ($\%$)}                         & B@1  & CIDEr & Acc ($\%$)             & \multicolumn{1}{c}{}                          \\ \hline
X-Decoder (our impl.)             & original & \multicolumn{1}{c|}{37.7}                   & --  & -- & -- & -- & -- & --& -- & --    & 57.8 & 46.8  & \multicolumn{1}{c|}{--}  &  \multicolumn{1}{c}{62M}                                               \\
X-Decoder (our impl.)            & multi-head                                        & \multicolumn{1}{c|}{38.1}                   & 0.432                                  &89.6 & 68.3 & 80.5 & 84.4 & 73.0& 77.1 & 79.4    & 59.5 & 50.0  & \multicolumn{1}{c|}{--}  &  \multicolumn{1}{c}{63M}                                               \\ \hline
\rowcolor{gray!15}Ours                      & task-prompt                                       & \multicolumn{1}{c|}{\textbf{38.5}}                           &\textbf{0.425}      &\textbf{90.2}      &\textbf{68.7}      &\textbf{81.1}     &\textbf{84.9}       &\textbf{73.8}      &\textbf{77.8}      &\textbf{80.1}     &\textbf{61.0}     &\textbf{52.5}      & \multicolumn{1}{c|}{\textbf{53.7}}  &  \multicolumn{1}{c}{62M}                                              \\ \hline
\end{tabular}%
}
\caption{\textbf{Ablation study of task-specific prompt} in \textsc{@Model}. ``our impl.'' means our implement based on the original paper, ``multi-head'' means we add multiple output heads (a 3-layer MLP for each task) to the original model to achieve different tasks on \textsc{@Bench}.}
% \vspace{-1em}
\label{ab:prompt}
\end{table*}

\subsection{Efficiency-Performance Trade-off}
\textsc{@Model} exhibits a favorable balance between efficiency and performance. As demonstrated in Table~\ref{exp:multi_task}, \textsc{@Model} and the small version of Unified-IO have a similar number of parameters, yet \textsc{@Model} outperforms Unified-IO (S) in all tasks they have in common, and surpass the basic version of Unified-IO. In Table~\ref{exp:single_task}, with almost the same number of parameters, the performance of \textsc{@Model} is better than many single-task SoTA models without pre-training. This further demonstrates that \textsc{@Model} can achieve the SoTA level on multiple tasks with fewer parameters, which is crucial for ATs with limited computational capacity.

\subsection{Ablation Study}
To dive deep into the model design and to investigate the effect of the proposed method, we conduct ablation studies of the task-specific prompt, the tokenizer and vocabularies. 

\noindent\textbf{From Task-specific Prompt to Unified Input.\label{sec:unified-input}} 
As shown in Table~\ref{ab:prompt}, the ``image + prompt'' training paradigm is demonstrated to be more concise and compatible with a wider range of tasks, outperforming multi-output heads training across all tasks. While there isn't a significant difference in the number of parameters between the two approaches, one potential advantage of prompt-based training is its ability to use prompts to differentiate between various tasks during training earlier. This enables the model to extract distinct features for each task, leading to varied and improved results. During multi-output head training, models extract mixed features, and only the final output head is used to decode corresponding features for different tasks.
Besides, advanced output heads should be added when making the models with a multi-output head paradigm compatible with more tasks.
Therefore, for generalist models, the prompt-based training paradigm is more promising.

\noindent\textbf{Different Tokenizers and Vocabularies for OCR.\label{sec:ch_OCR}} 
As shown in Table~\ref{ab:OCR}, utilizing a character-based tokenizer can lead to a performance improvement of $4.3\%$, and incorporating a limited vocabulary can further enhance performance by $3.3\%$. 
OCR solely recognizes image information, while captioning and VQA require processing and logical reasoning of image information to obtain reasonable answers. Therefore, when integrating tasks at different granularities, such as ``relatively simple'' OCR and ``more complex'' captioning into one model, it is suggested to carefully select different tokenizers and vocabularies to achieve the best performance on each task. 

\begin{table}[t]
\renewcommand\arraystretch{2}
\setlength{\tabcolsep}{2pt}
\resizebox{\columnwidth}{!}{%
\begin{tabular}{cc|cc|ccccccc}
\hline
\multicolumn{2}{c|}{\textbf{Tokenizer}} & \multicolumn{2}{c|}{\textbf{Vocabulary}} & IC13 & IC15 & SVT & IIIT5K & SVTP & CUTE & Avg. \\
sub            & ch            & complete        & limited       & \multicolumn{7}{c}{Acc ($\%$)}                        \\ \hline
\checkmark               &               & \checkmark                &               &95.1      &73.5      &90.1     &82.2        &84.5      &81.6      &82.4     \\
               & \checkmark              & \checkmark                &               &95.8      &80.0      &90.0     &87.5        &85.0      &90.0      &86.7     \\
               & \checkmark              &                 & \checkmark              & \textbf{97.1}     & \textbf{84.4}     & \textbf{92.6}    & \textbf{90.3}       & \textbf{88.7}     & \textbf{93.1}     & \textbf{90.0}     \\ \hline
\end{tabular}%
}
\caption{\textbf{Ablation study of tokenizer and vocabulary} for task text recognition. ``sub'' and ``ch'' denote subword-based and character-based, respectively.}
% \vspace{-1em}
\label{ab:OCR}
\end{table}

\section{Discussion\label{discussion}}
\label{future_work}
\noindent \textbf{Qualitative Analysis of User Study.} 
Guided by the human-in-the-loop user study, we create the benchmark \textsc{@Bench}. Score potential functions from multiple perspectives to ensure the selected functions and tasks are reasonable and important to PVIs. The five tasks in \textsc{@Bench} are closely related to these functions and are one of the ways to achieve these functions. Furthermore, based on these tasks implemented by \textsc{@Model}, we are prepared to carry out further assistive function development in the future, such as indoor obstacle avoidance, text recognition,  common objects detection, initial understanding of unfamiliar scenes, \textit{etc}.

Additionally, we analyze the comments given by the blind user and list a few. About OCR, \textit{``Usability by completely blind people would be very important to me. Existing systems of this type have difficulty selecting the right target. Text recognition on a document in front of me or on packaging in my hand is also very possible with a smartphone. However, if the function is able to read door signs, hanging posters or street signs that are not within direct reach, that would be an extremely useful function."}. According to the scores in the questionnaire and comments, PVIs attach great importance to text recognition, especially in some special scenarios. Therefore, unlike other generalist models that ignore OCR task, \textsc{@Bench} contains OCR task and introduce multiple OCR datasets, covering various text scenarios. About object recognition, \textit{``This is a function I would definitely expect!", ``It would be important, on the one hand, to have a high level of reliability of recognition and, on the other hand, to be able to determine, even for completely blind people, that the correct object is being recognized."}. By introducing multi-category ADE20K and combining panoptic segmentation, \textsc{@Model} can recognize a large number of stuffs and things, compared with detection, semantic segmentation. At the same time, the more categories of objects are recognized, the accuracy of recognition will increase and the recognition will be more reliable. In sum, these positive comments and great suggestions inspire our work and provide guidance for future work of exploring VLMs for assistive technology.

\noindent \textbf{Future Directions.} 
The extensive quantitative and qualitative results have demonstrated the strong effectiveness and efficiency of our \textsc{@Model} for a variety of assistance-related tasks at different granularities. Upon the current, we see two directions worth future explorations: (1) \textit{Pre-training}. Currently, 
we did not perform pre-training and 
\textsc{@Model} can reach a level close to the pre-trained SoTA. We believe that after pre-training, \textsc{@Model} can achieve higher performance. (2) \textit{Functions development and deployment}. Going back to the user study, we came up with the idea for work precisely because we understood the difficulties that PVIs encounter in daily life. Existing assistive systems generally can only implement one or a few functions.
A future work is to implement a multi-functional assistive system based on \textsc{@Model}. 
More discussions locate in the supplementary material. 

\section{Conclusion\label{conclusion}}
In this work, we introduce \textsc{@Bench}, a multi-modal, multi-task, multi-dimension benchmark for the evaluation of generalist VLMs that can empower assistive technology and help PVIs. Based on the human-in-the-loop user study with the target group, our \textsc{@Bench} not only considers 5 practical tasks closely related to the daily lives of PVIs, but also takes into account the efficiency guideline for VLMs.
Furthermore, we present a unified and multi-task \textsc{@Model} to address the multiple vision-language tasks. Thanks to the unified task-specific prompt design, our model can use one suit of parameters to address all 5 tasks and achieve competitive results. Extensive experiments and qualitative analysis prove the effectiveness of the proposed \textsc{@Bench} and \textsc{@Model} in helping PVIs.
We hope this work can provide inspiration for the design of next-generation assistive systems for helping PVIs.

\noindent {\\ \textbf{Acknowledgment.}
This work was supported in part by the Ministry of Science, Research and the Arts of Baden-Wurttemberg (MWK) through the Cooperative Graduate School Accessibility through AI-based Assistive Technology (KATE) under Grant BW6-03, in part by the Federal Ministry of Education and Research (BMBF) through a fellowship within the IFI program of the German Academic Exchange Service (DAAD), and in part by Future Mobility Grants from InnovationCampus Future Mobility (ICM). 
We thank HoreKA@KIT, HAICORE@KIT, and bwHPC supercomputer partitions.
}

%%%%%%%%% REFERENCES
{\small
\bibliographystyle{ieee_fullname}
\bibliography{main}
}

\clearpage
\appendix

\section{Model Architecture}
The overall architecture of \textsc{@Model} is a generic encoder-decoder design as shown in main paper. We follow X-Decoder~\cite{zou2023generalized} to adapt Focal-T~\cite{yang2022focal} as image encoder $\mathbf{Enc_{I}}$ and use a number of transformer layers as text encoder $\mathbf{Enc_{T}}$. Decoder is a common Transformer~\cite{zou2023generalized} decoder structure with self- and cross-attention layers.
\subsection{Formulation}
First, we use image encoder $\mathbf{Enc_{I}}$ to extract multi-scale features $\mathbf{Z}$ from input image $\mathbf{I}{\in}\mathcal{R}^{H \times W \times 3}$:
\begin{equation}
\mathbf{Z} = \mathbf{Enc_{I}(I)} = \left \langle {\mathbf{z}_l} \right \rangle_{l=1}^{L}
 \label{eq:image_encode}
\end{equation}
where $\mathbf{z}_l \in \mathcal{R}^{H_l \times W_l \times d}$ and $\{ H_l, W_l\}$
is the size of feature map at level $l$ and $d$ is the feature dimension. Then, we use the text encoder $\mathbf{Enc_{T}}$ to encode a task-specific prompt into $\textbf{P}=	\left \langle {p_1}, \cdot\cdot\cdot, {p_n}\right \rangle$ of length $n$. Afterwards, we use the same text encoder $\mathbf{Enc_{T}}$ to encode a textual label into $\mathbf{Q^t} = \left \langle {q_1}^t, \cdot\cdot\cdot, {q_n}^t \right \rangle$ and create a latent queries $\mathbf{Q^l} = \left \langle {q_1}^l, \cdot\cdot\cdot, {q_m}^l \right \rangle$ as inputs of decoder. All these features are fed into \textsc{@Model} to predict the outputs:
\begin{equation}
    \left \langle \mathbf{O^p}, \mathbf{O^s} \right \rangle = \mathbf{@Model}(\left \langle \mathbf{P}, \mathbf{Z}\right \rangle; \left \langle \mathbf{Q^l}, \mathbf{Q^t}\right \rangle),
 \label{eq:all_task}
\end{equation}
where $\mathbf{O^p}$ and $\mathbf{O^s}$ are the pixel-level outputs and token-level
semantic outputs, respectively.

\subsection{Tasks}
Based on the aforementioned designs, \textsc{@Model} can be effectively employed to integrate various vision and vision-language tasks by utilizing different input combinations.

\noindent\textbf{Pixel-level Output Tasks. }
For these tasks, such as panoptic segmentation and depth estimation, there is no textual
label as input for decoder:
\begin{equation}
 \mathbf{O^p} = \mathbf{@Model}(\left \langle \mathbf{P}, \mathbf{Z}\right \rangle; \mathbf{Q^l}),
 \label{eq:pixel}
\end{equation}
where $\mathbf{O^p}$ has the same size of $\mathbf{Q^l}$.

\noindent\textbf{Token-level Output Tasks.}
For OCR, captioning and VQA, they require both latent and text queries as inputs. Hence, Eq.~(\ref{eq:all_task}) is adapted to:
\begin{equation}
 \mathbf{O^s} = \mathbf{@Model}(\left \langle \mathbf{P}, \mathbf{Z}\right \rangle;  \left \langle \mathbf{Q^l}, \mathbf{Q^t}\right \rangle),
\end{equation}
where $\mathbf{O^s}$ correspondingly has equal size of $\mathbf{Q^t}$, and no pixel-level output are predicted. All predictions follow an auto-regressive strategy.

\section{Loss Functions}
\subsection{Pixel-level Output Loss}
\noindent\textbf{Segmentation Loss.} There are two losses on the segmentation corresponding to two tasks. For mask classification, we use text encoder $\mathbf{Enc_{T}}$ to encode all $N$ class names including ``background" into $N$ text embeddings $\mathbf{E}_{cls} {\in} \mathcal{R}^{N \times C}$ and take it to represent the concept. Afterward, we take the first ($m-1$) latent queries and compute the dot-product between these outputs and concept embeddings to obtain an affinity matrix $\mathbf{S}_{cls}{\in}\mathcal{R}^{(m-1) {\times} N}$ and compute $\mathcal{L}_{cls} {=} \mathbf{CE}(\mathbf{S}_{cls}, \mathbf{y}_{cls})$, with the ground-truth class $\mathbf{y}_{cls}$. For mask prediction, we use Hungarian matching~\cite{carion2020end,cheng2022masked} to find the matched entries of first ($m-1$) outputs to ground-truth annotations.  Afterward, we to use binary cross-entropy loss $\mathcal{L}_{bce}$ and dice loss $\mathcal{L}_{dice}$ to compute the loss. Thus, the overall training loss function of panoptic segmentation is: 
\begin{equation}
\mathcal{L}_{ps} = \lambda_{cls}\mathcal{L}_{cls} + \lambda_{bce}\mathcal{L}_{bce} + \lambda_{dice}\mathcal{L}_{dice},
\label{eq:ps}
\end{equation}
where $\lambda_{cls}$, $\lambda_{bce}$ and $\lambda_{dice}$ are coefficient weights to control different losses

\noindent\textbf{Depth Estimation Loss.} Given  the prediction $\mathbf{O^p}$ derived from $m$ latent queries, we use the last ($m$-th) latent querie to make depth prediction. In order to calculate the distance between predicted output $\mathbf{\widehat{Y}}_{de}$ and ground truth $\mathbf{{Y}}_{de}$, we use scale-invariant log scale loss~\cite{eigen2014depth,kim2022global}. The equation of training loss is as follows:
\begin{equation}
\mathcal{L}_{de} = \frac{1}{n}\sum_{i}{d_i}^2 - \frac{1}{2}(\frac{1}{n}\sum_{i}{d_i})^2,
\label{eq:de}
\end{equation}
where $d_i = log(y_i) - log(\widehat{y_i})$, $y_i$ and $\widehat{y_i}$ are $i$th pixel-value of $\mathbf{{Y}}_{de}$ and $\mathbf{\widehat{Y}}_{de}$, respectively.

\subsection{Token-level Output Loss}
For token-level tasks, we begin by extracting embeddings for all tokens in the vocabulary, which has a size of $V$, from the text encoder. Using the last $n$ semantic token-level outputs from \textsc{@Model}, we calculate the dot product with all token embeddings to generate an affinity matrix $\mathbf{S}_{token}{\in}\mathcal{R}^{n \times V}$. Subsequently, we compute the cross-entropy loss $\mathcal{L}_{token} {=} \mathbf{CE}(\mathbf{S}_{token}, \mathbf{y}_{token})$, where $\mathbf{y}_{token}$ represents the ground-truth next-token id.

\subsection{Multi-task Training Loss}
During multi-task training, we calculate losses on the top decoder layers for each task to guide the model to converge faster in the early training stage and accelerate the overall training process. 
The overall training loss function is:
\begin{equation}
    \sum_{task\in\{ps, de, ocr, ic, vqa\}}\sum_{i=1}^{nl_{task}}\lambda_{task}\mathcal{L}_{task},
\end{equation}
where $nl_{task}$ represents the number of decoder layers that need to calculate the loss for different task, $\lambda_{task}$ and $\mathcal{L}_{task}$ are loss weights and losses for different task, respectively.

\begin{figure}[t]
  \centering
   \includegraphics[width=1.0\linewidth]{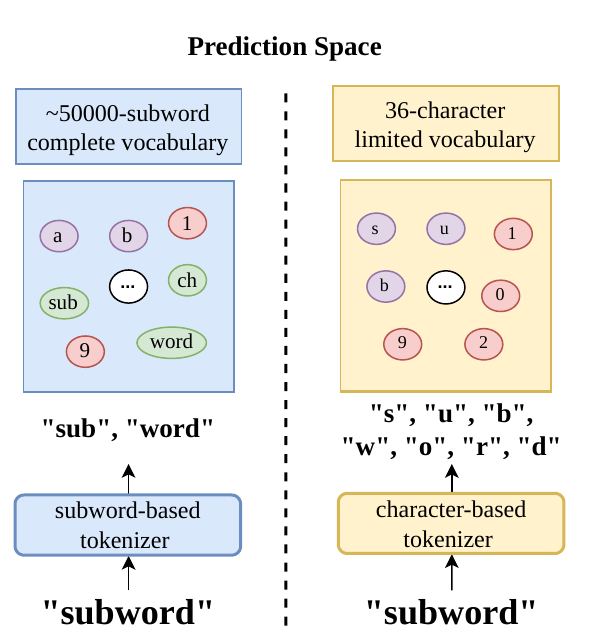}
   \caption{Comparison between subword-based tokenizer and character-based word tokenizer in our proposed \textsc{@Model}}
   \vspace{10mm}
   \label{fig:ocr_predict}
\end{figure}

\section{Implementation Details}

\subsection{Multi-task Training}
\noindent\textbf{Training Setting.} Since the number of images in OCR training dataset is much larger than datasets for the other tasks, we define OCR training dataset as the major dataset for multi-task training. It means that the total number of iterations is calculated based on the number of images in the OCR dataset. The batch sizes for panoptic segmentation, depth estimation, OCR, captioning, and VQA are 4, 4, 768, 8, and 4, respectively, to accommodate datasets of different sizes for various tasks. The model is trained with 15 epochs based on OCR datasets on 4 A100 (40G). The AdamW optimizer is used with the initial learning rate 1$^e$-5. A step-wise scheduler is used to decay the learning rate by 0.1 on the fraction [0.6, 0.8] of training steps.

\noindent\textbf{Hyperparameter Choice.} In \textsc{@Model}, the decoder has 7 decoder layers. Due to segmentation and OCR are the top two scored tasks (from user study), and the OCR datasets are very large, the amount of data for each task is unbalanced, we set $nl_{task}$ as 6, 3, 6, 3 and 3 for panoptic segmentation, depth estimation, OCR, captioning and VQA, respectively, to allow the model to focus more on segmentation and OCR during training. We set loss weights $\lambda_{task}$ as 1, 10, 10, 2 and 2, respectively. Because the loss values of OCR and depth estimation in the later stage of training are very small, in order to minimize significant differences in the loss magnitude for each task as much as possible, we have made such a setting. And in Eq.~(\ref{eq:ps}), we set $\lambda_{cls}$ = 2, $\lambda_{bce}$ = 5 and $\lambda_{dice}$ = 5 as default, following the settings of X-Decoder.

\subsection{Single-task Training}
All tasks are trained with AdamW as the optimizer on 4 1080Ti (11G), except OCR. The initial learning rate is 1$^e$-5 and reduced by 10 times after 60\% and 80\%.

\noindent\textbf{Panoptic Segmentation.}
We train the model for 50 epochs. We set the image resolution to 640${\times}$640 and the batch size to 4.

\noindent\textbf{Depth Estimation.}
We train the model for 50 epochs. We set the image resolution to 480${\times}$640 and the batch size to 16. Note that this task is very unstable and requires careful hyperparameter tuning. If you encounter training errors, you can increase the batch size, reduce the learning rate or training with single precision (FP32).

\noindent\textbf{Optical Character Recognition.} 
We train the model for 10 epochs on 4 A100 (40G). We set the image resolution to 64${\times}$200 and the batch size to 1024. 

\noindent\textbf{Image Captioning.} 
We train the model for 50 epochs. We set the image resolution to 480${\times}$640 and the batch size to 16. We use all captions for training and do not use beam search and CIDEr optimization.

\noindent\textbf{Visual Question Answering.}
We train the model for 30 epochs. We set the image resolution to 480${\times}$640 and the batch size to 16.

\subsection{Character-based Tokenizer with Limited Vocabulary for OCR}
In our main paper, we observed that subword-based tokenizer with complete vocabulary hurts the performance of OCR task. In Fig.~\ref{fig:ocr_predict}, we show how to use character-based tokenizer and much smaller limited vocabulary to perform OCR task. Using a character-based word tokenizer to divide the text that needs to be recognized into characters one by one, model only needs to predict token from the limited vocabulary space, and do not need to select candidate subword from the complete vocabulary. This reduces the prediction space and improves the accuracy of prediction.

\section{User Study}
\subsection{Comments on Generalist Assistance Systems}
By conducting the questionnaire survey, we communicated with visually impaired individuals to comprehend the functionalities they expect a generalist assistive system should possess. We got some thoughts like: \textit{``It should find the door, look for stairs in an open area, read the house/room number, read signs/plates, describe the environment, warn me of obstacles, and can navigate the corridors with a floor plan."}. Some participants also described specific usage scenarios, \textit{``I would use navigation and obstacle detection systems outside. It should warn me of obstacles or describe something I'm about to encounter. For example, if I'm navigating outside and there's a road ahead, then it should say if it has a roundabout or an intersection. Or, if there is a railroad crossing, announce something similar. It would be cool if there was an all around view. The system says, the front of you is street and the back is a building, left is bike racks, etc. If there is a name of the store, read it out. The most important thing is to have a general navigation ability based on this all around view. If I then say navigate to the store (name of the store) recognized by this system, then it should navigate me there."}. Based on these thoughts and comments, essential functions identified by People with Visual Impairments (PVIs) that a generalist assistive system should include are: 
\begin{compactitem}
\item[(1)] \textbf{Navigation and Obstacle Avoidance.} A critical component is a navigation system integrated with obstacle detection capabilities. PVIs desire a system that allow for interactive navigation, where users can request directions to specific locations identified by the system.

\item[(2)] \textbf{Text Recognition and Environmental Description.} The ability to recognize and verbally relay textual information is also important. This includes identifying and reading door labels, room numbers, and signs. Furthermore, recognizing the names of stores, significant landmarks or other text contributes to better environmental understanding and orientation.

\item[(3)] \textbf{Comprehensive Scene Interpretation.}  PVIs expressed a desire for a system that provides a holistic view of their surroundings. This ``all-around vision" function should describe streets, buildings, and other elements in the vicinity. 

\item[(4)] \textbf{Integration of Text-to-Speech Technology.} Incorporating text-to-speech technology for dynamic interaction is also valuable. 
\end{compactitem}

\subsection{More Comments}
\noindent\textbf{Navigation.} The majority of participants (5 P: 5 participants) prioritize outdoor navigation, noting its greater complexity and risk. They highlight that outdoor environments pose larger obstacles, longer and more complex routes, and a higher likelihood of getting lost compared to indoor scenarios. One participant emphasized, \textit{``Outdoor navigation is much more important. Indoors, the reach of a cane is much more likely to adequately capture the surroundings. The distances are shorter and the density of people is higher."}. Another added, \textit{``Definitely outdoors. If I have to go into a building I don't know, it will probably only be for once. It's not worth learning a way to do that."}. The unpredictability of outdoor spaces, such as traffic, was also mentioned as a significant factor. Conversely, a minority (2 P) believes that indoor navigation is more important. They mention the challenges of navigating within large unfamiliar buildings, locating specific rooms, stairs, elevators, or exits, walking across a large open-area, and walking in rooms with highly differentiated structures, such as restrooms. Importantly, they spend most of their time indoors.

\noindent\textbf{Text Recognition.} Today, PVIs mainly use screen readers to recognize digital texts and usually use smartphones, or smartphones Apps to read non-digital texts. However, they find non-digital text reading is difficult and cumbersome, like \textit{``Everything I receive on paper in the post annoys me. I use apps like Seeing AI and Be My Eyes or the iPhone's magnifying glass to read non-digital texts, but using a smart glass to read these text directly would be better."}. They also pointed out that it is also important for them to read signs to find the right floor or hallway and read door numbers to enter the right room.

\noindent\textbf{Other Functions.} About depth estimation, \textit{``This function helps one develop a mental map of an environment. You get the proportions well."}. About object location, \textit{``In my personal environment I am always very sure where all the things I am looking for are. However, locating a true one in a larger shelf section of 3-4 meters would be very useful. A function that detects objects that don't belong in that space would also be very helpful to check a room for overlooked clutter. The glasses could use a reference photo of the tidy room and then report any anomalies, such as dirty dishes on the table or socks on the floor."} and \textit{``If I only need it if I can't find something in my apartment, it could make the search easier, but I would need it pretty rarely."}. About surroundings understanding, \textit{``It would be important to me that the description be highly efficient. The short form is always first"} and \textit{``Most of the time we are not interested in the scene because it is too much information for us. But descriptions of photos, environments, etc. are very exciting. ChatGPT is really great."}. About scene recognition, \textit{``Perhaps interesting for recognize different scenarios, but a correspondingly efficient description of the image would serve the same purpose. I can't imagine a situation where I would need room detection. I usually know which room I'm going to or being led into."}. About visual Q\&A, \textit{``This function would make it possible to expand a short initial description of an image dynamically and according to your own needs. That would improve the overall function enormously."}.
 
\noindent\textbf{Interaction.} If there were such a general system, PVIs prefer interacting with system through discrete button presses or subtle gestures (6 P), rather than voice commands (1 P) for privacy reasons, when inputting instructions. For receiving system feedback, they show a preference for auditory feedback (for general purpose) and vibrations (for special purpose such as obstacle avoidance).

Based on these comments and ideas, it becomes evident that for PVIs, navigation and quick, direct recognition of non-digital text are the two most critical functionalities. Meanwhile, the multifaceted nature of navigation encompasses functions like environmental comprehension, obstacle avoidance, path planning, voice guidance and \textit{etc.} These insights serve as valuable guidance for our work. Furthermore, the analysis of participants' relevant feedback has provided us with an initial understanding of creating a universal assistive system.
\begin{figure*}[!t]
  \centering
   \includegraphics[width=1.0\textwidth]{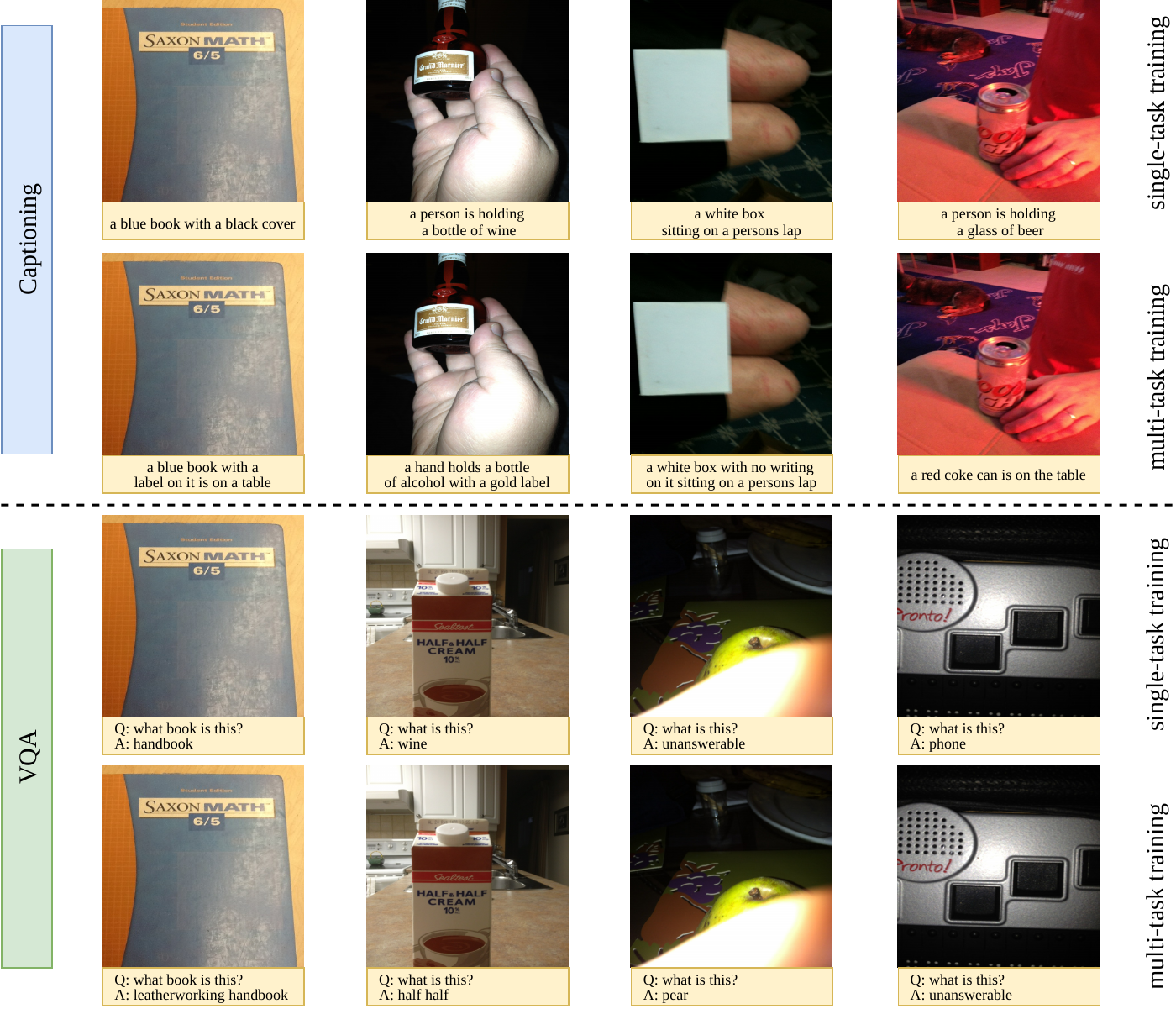}
   \caption{Examples to show the promotion of vision for vision-language and complementariness between different vision-language tasks.}
   \label{fig:comp_vis}
\end{figure*}

\begin{table}[t]
\centering
\renewcommand\arraystretch{1.0}
\setlength{\tabcolsep}{4pt}
\resizebox{\linewidth}{!}{%
\begin{tabular}{ccc|c|cc|ccccc}
\hline
\multicolumn{3}{c|}{Task} & \multicolumn{1}{c|}{ADE-150} & \multicolumn{2}{c|}{VizWiz\_Cap} & \multicolumn{5}{c}{VizWiz\_VQA} \\
PS            & IC          & VQA  &PQ &B@1 &CIDEr &Other & Unans & Yes/No&Number &Acc($\%$)                                \\ \hline
\checkmark               &               &                &39.2 &-- &-- &--  &-- &-- &-- &--       \\
             &\checkmark                &               &-- &60.0 &45.1 &--  &-- &-- &-- &--       \\
             &               & \checkmark               &-- &-- &-- &30.5  &92.1 &70.1&13.7 &49.1       \\
\checkmark              & \checkmark                &               &37.7 &57.8 &46.8 &--   &-- &-- &-- &--      \\
              &\checkmark                 & \checkmark              &-- &59.8 &46.3 &32.2   &86.5 &73.4 &16.4&48.8      \\ 
\checkmark              &\checkmark                  & \checkmark              &38.5 &61.0 &52.5 &39.4   &88.2 &70.1 &10.8&53.7      \\ 
\hline
\end{tabular}%
}
\caption{\textbf{Comparison of results of mixed training for different tasks.} Note:``Other", ``Unanswerable", ``Yes/No", ``Number" are 4 different answer types for VQA. (PS = panoptic segmentation, IC = image captioning).}
\label{tb:mix_training}
\end{table}

\begin{figure*}[!t]
  \centering
   \includegraphics[width=1.0\textwidth]{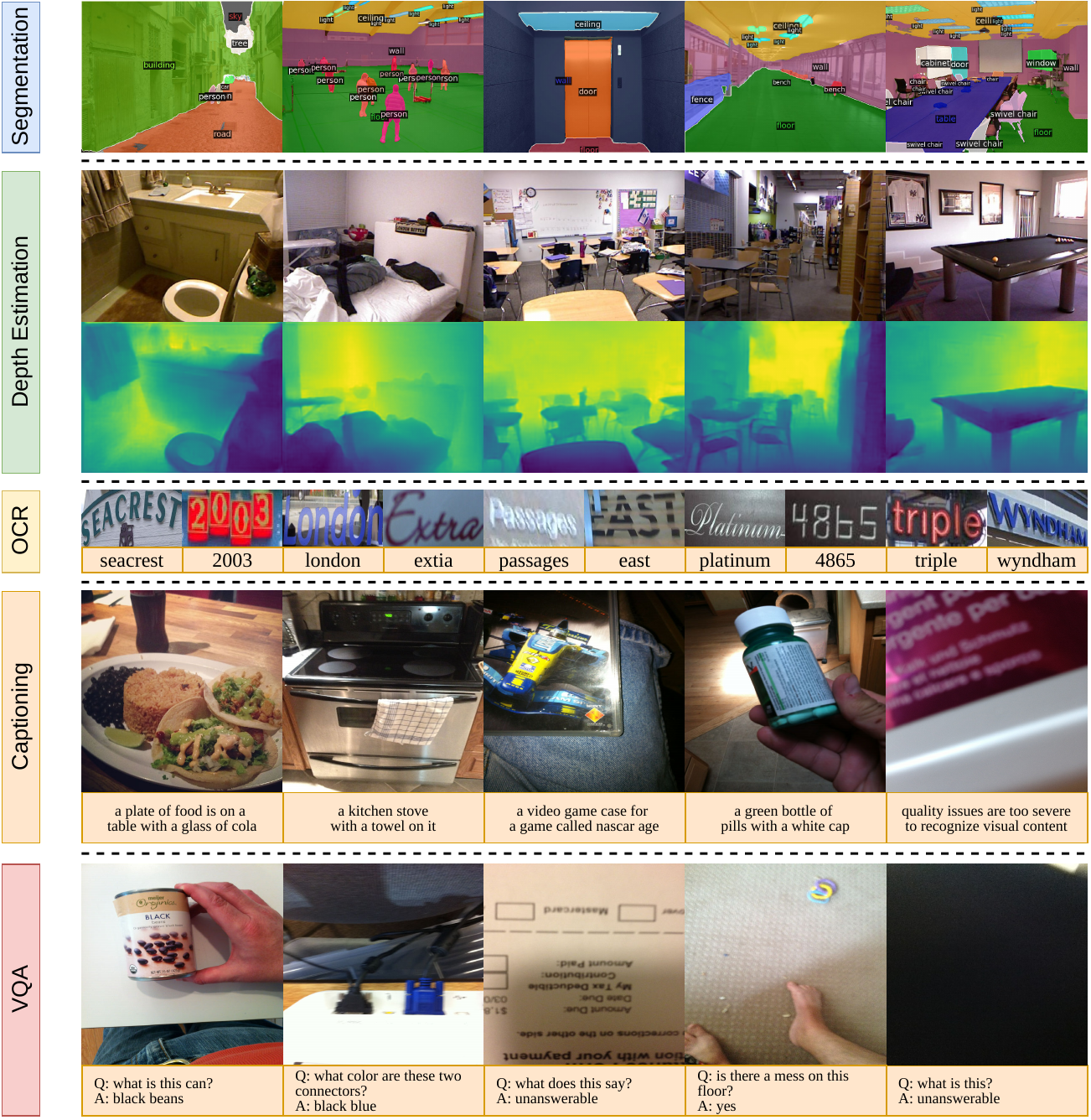}
   \caption{Examples on different test datasets. These images cover a diversity of visual domains and concepts in the daily life of PVIs.}
   \vspace{2mm}
   \label{fig:test_vis}
\end{figure*}

\begin{figure*}[!t]
  \centering
   \includegraphics[width=1.0\textwidth]{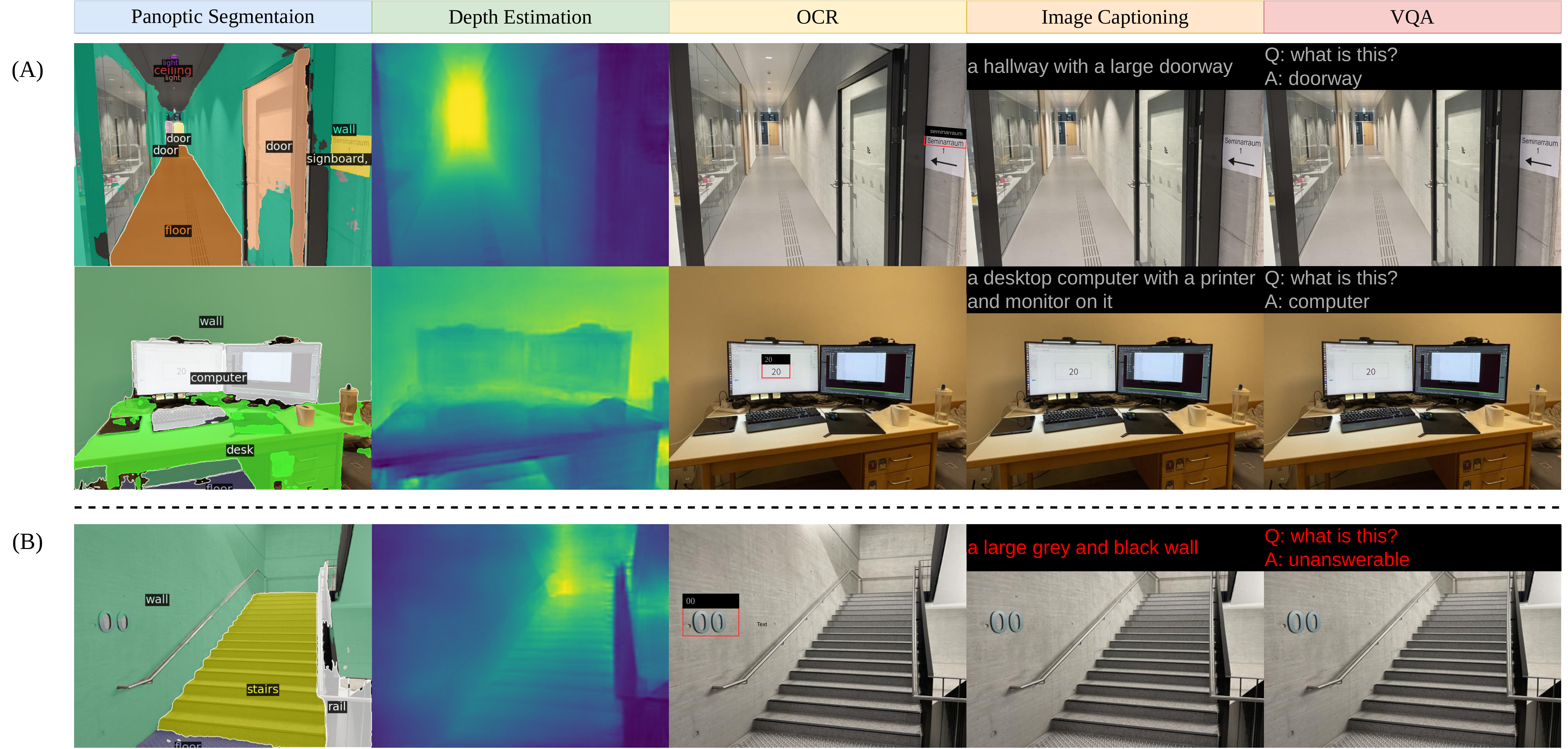}
   \caption{Examples on real-world scenes. These images were randomly collected by using mobile phone.}
   \vspace{2mm}
   \label{fig:zero_shot}
\end{figure*}

\section{More Experiments}
\subsection{Complementariness in Multitasking}
As shown in the experiments section, our \textsc{@Model} exhibits a strong performance in captioning and VQA under multi-task training. Here, we further study the role of segmentation objectives in vision-language (VL) understanding, as well as the role of different vision-language understanding tasks on each other. To investigate, we mix different tasks for training. 
In Table~\ref{tb:mix_training}, for captioning, when jointly trained with VQA or PS, or all tasks, CIDEr improved by $1.2$, $1.7$ and $7.4$ respectively. For VQA, we report 5 numbers for better analysis, namely the accuracy for 4 Q\&A types: \textit{other}, \textit{unanswerable}, \textit{yes/no}, \textit{number}, and the overall accuracy. From the comparison of these numbers, when training VQA alone, the model tends to predict ``unanswerable" to improve the accuracy. Because in the dataset, the \textit{unanswerable} type of Q\&A is the most common. For other types of Q\&A, the accuracy is relatively lower because a deeper or more granular understanding of the semantic information of image is required to predict the correct answers. After joint training with captioning, the accuracy of \textit{unanswerable} type Q\&A decreased, and the accuracy of other types increased. The model does not just return ``unanswerable" blindly but understands more semantic information of the image and then make predictions. When all tasks are trained together, the accuracy of \textit{other} type Q\&A is greatly improved ($+8.9\%$). We analyze that it is because the question of this type of Q\&A is usually ``\textit{what is this?}'', and the segmentation task naturally has a very good assisting effect in answering this question. Segmentation data can help models to learn more fine-grained visual understanding and consequently benefit vision-language tasks. We also give some examples to show these improvements in Fig.~\ref{fig:comp_vis}. Along with our findings in the main paper, we conclude that segmentation has clear benefits to VL learning and different VL tasks are complementary to each other.

\section{More Visualization}

\subsection{Visualization on Test Datasets}
We present a comprehensive visualization of our model's performance on the test datasets in Fig.~\ref{fig:test_vis}. For segmentation, we show some results in outdoor scene, indoor scene, multi-person scene, especially the open-area mentioned by the PVIs. For OCR, various types of text recognition results can show the robustness and generalization of \textsc{@Model}. For other task, \textsc{@Model} can also perform well.

\subsection{Zero Shot}
Finally, we apply the 5 tasks in a zero-shot manner to show the generalization ability of \textsc{@Model}. \textsc{@Model} performs well on three tasks: segmentation, depth estimation, and OCR, as shown in Fig.~\ref{fig:zero_shot}. However, for open-ended tasks, captioning and VQA , the performance on out-of-dataset data can sometimes be less satisfactory (Fig.~\ref{fig:zero_shot} (B)). Therefore, it may be necessary to perform large-scale pre-training to enhance the model's capability for handling these tasks well in zero-shot.

\section{More Discussion}
This section discusses the limitations and future work of this work for more insights on the research in this track.

\noindent\textbf{Pre-training.}
In the main paper, we did not perform pre-training. This has a certain impact on the capability of zero shot, especially for open-ended tasks. In the future, we plan to conduct pre-training on large-scale corpora to enhance the model's zero-shot capability. Additionally, we use a unified language encoder to encode text in \textsc{@Model}. Pre-training can enrich the vocabulary size, thereby improving the model's ability to open-vocabulary segmentation. The importance of this open-vocabulary capability for practical applications is self-evident, especially for blind users. As mentioned by blind users in user study, they require systems with high object recognition accuracy. When the model has seen a greater variety of objects and can distinguish between them, the recognition accuracy also increases. Additionally, this open-vocabulary capability allows the model to handle previously unseen objects. In sum, after pre-training, the model can better handle the diversity, complexity and unpredictability of usage scenarios.

\noindent\textbf{Multi-task Training.}
As shown in the main submission, \textsc{@Model} performs well on the OCR task during single-task training, but there is a certain gap in performance during multi-task training. Our analysis suggests that the OCR dataset is too large, and the model does not balance multiple tasks during training. When dealing with multi-task training with extremely imbalanced dataset sizes, it is not enough to merely adjust loss weights differently. In the future, we may try more optimization methods for multi-task learning to ensure performance without greatly increasing the training time.

\noindent\textbf{Functions Development and Model Deployment.}
In our user study, we have identified several potential and crucial functions that received unanimous agreement from participants. Furthermore, it's important to note that \textsc{@Model} is not limited to these five tasks alone; it can be extended to more uni-modal or multi-modal tasks to provide more functionalities. Our future research direction will focus on building a PVIs-Centred generalist assistive system, leveraging \textsc{@Bench} and \textsc{@Model} as cornerstones, to develop a wide range of practical functions and services.
As for model deployment, although \textsc{@Model} achieves high performance on multiple datasets, since the model is based on Transformer, its costs are larger than the non-Transformer models. 
Additionally, though \textsc{@Model} only has 62M parameters, it is still difficult to deploy such a model in the portable device used by PVIs. Therefore, in our future work we will discover how to extract or compress \textsc{@Model}  into an efficient light-weight model.

\end{document}